# Attention Visualizer Package: Revealing Word Importance for Deeper Insight into Encoder-Only Transformer Models

Ala Alam Falaki[1*] and Robin Gras[1]

[1] *School of Computer Science, University of Windsor. Windsor, Ontario, Canada. N9B3P4*

* Corresponding Author. Email: alamfal@uwindsor.ca

## Abstract

This report introduces the "Attention Visualizer" package, which is crafted to visually illustrate the significance of individual words in encoder-only transformer-based models. In contrast to other methods that center on tokens and self-attention scores, our approach will examine the words and their impact on the final embedding representation. Libraries like this play a crucial role in enhancing the interpretability and explainability of neural networks. They offer the opportunity to illuminate their internal mechanisms, providing a better understanding of how they operate and can be enhanced. You can access the code and review examples on the following GitHub repository: https://github.com/AlaFalaki/AttentionVisualizer.

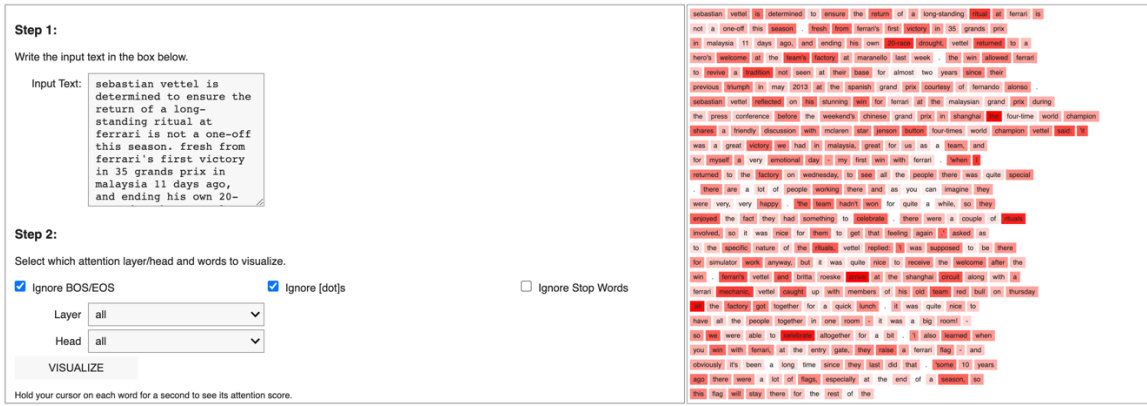

*Fig 1. The library's user interface operated within a Google Colab instance. (left) The input fields and settings to calculate the attention scores. (right) The visualization takes the form of a heatmap, with darker shades of red indicating higher scores.*

## 1. Introduction

From the time of their introduction, neural networks were regarded as enigmatic systems, often resembling black boxes (Castelvecchi 2016) due to the difficulty in grasping their underlying mechanisms. Nevertheless, in recent years, numerous researchers have uncovered the inner workings of these networks and clarified the factors contributing to their functionality. Various approaches exist to assist in enhancing the interpretability of different types of neural networks.

There are research studies that have directed their focus towards different facets of these models, as evident in the paper on extracting rules from the network (Arbatli and Akin 1997), which emphasizes exploitability. Additionally, the xNN network from (Yang, Zhang, and Sudjianto 2019) is designed with architecture constraints to make the networks more interpretable. Another area of research pertains to the visualization of these models. Existing literature concentrates on visualizing the attention mechanism within vision-related tasks. (Yeh et al. 2023) Furthermore, the BertViz (Vig 2019) library focus on visualizing the self-attention scores in natural language processing, and more specifically, the BERT (Devlin et al. 2018) model.

Visualizing neural networks holds significant importance as it provides a tangible means to demystify their core operations and enhance our understanding of their inner dynamics. These intricate networks, inspired by the human brain, consist of numerous layers and interconnected nodes, making them challenging to comprehend solely through abstract mathematical descriptions. Visualizations offer an intuitive way to represent the flow of information, the weights assigned to connections, and the transformations occurring at each layer.

By translating abstract concepts into visual elements, researchers, practitioners, and even non-experts can gain insights into how neural networks learn, generalize, and make predictions. Such visualizations foster transparency, allowing to detect potential biases, anomalies, or areas of improvement within the network's structure and performance. In essence, visualizing neural networks bridges the gap between their computational complexity and human comprehension, enabling more effective development, interpretation, and refinement of these powerful tools across various domains.

This report introduces the "Attention Visualizer" library, designed to visualize the significance of words based on the attention mechanism scores within transformer-based models. Subsequent sections of this paper will delve into the library's design decisions, features, and its practical applications.

## 2. Design and Architecture

As illustrated in Figure 1, the library showcases a user-friendly interface to input the text and obtain the attention score or the significance of each word in a form of a heatmap of the output. The precise score can be viewed by hovering the crosshair over the word. It is possible to customize the scoring criteria to align with diverse use cases through interaction with the controls, a topic we will explore further in the subsequent discussion.

The primary emphasis of this library lies in encoder-only models, aiming to enhance understanding of how Transformer-based models create embeddings representation for textual content. This constitutes the primary contrast distinguishing this work from similar visualization libraries. Their primary emphasis lies in utilizing the self-attention mechanism scores to illustrate token relationships. In contrast, our approach involves quantifying the contribution of individual words to the final representation.

In the upcoming subsections, the design decisions and challenges will be discussed to shed light on the details of our approach and provide comprehensive insights into its implementation.

### 2.1. Pre-Trained Model

As previously stated, the library is tailored for encoder-only models, with the default model chosen for this package being the base variant of the pre-trained RoBERTa (Liu et al. 2019) model. The Transformers library (Wolf et al. 2020) is employed to simplify the loading process and facilitate seamless integration of various models, enabling straightforward comparisons of their performance. One factor to consider when choosing a model is the predetermined maximum input size which is established during the pre-training phase. The RoBERTa model comes with a limitation of 512 tokens for input. Consequently, the library will truncate any text that exceeds this limit to the specified size.

### 2.2. Tokens vs Words

The tokenization process (Webster and Kit 1992) involves breaking down a text into smaller chunks, called tokens. These tokens can be words, subwords, or even characters, depending on the chosen tokenization strategy.

The tokens are stored within a dictionary and can be substituted with their corresponding IDs for input into the models. Although several techniques exist for tokenization, the Byte-Level BPE (Sennrich, Haddow, and Birch 2016) method stands out as a renowned approach that has demonstrated its effectiveness for transformer-based models. This method has the potential to divide a word into multiple tokens. For example, the word "tokenizing" would be represented as a combination of two tokens: "token" + "izing".

This introduces a certain level of complexity for visualization step. How do we address this, given that each token ("token," "izing") is assigned a score instead of the entire word? The solution implemented here involves selecting the token with the highest score as the score for the respective word. (For instance, if the initial token "token" has a score of 0.05 and the second token "izing" has a score of 0.01, the score of 0.05 will be chosen as the score for the word "tokenizing.")

## 2.3. Calculating the Score

The self-attention scores tensor possesses a size of [L×H×N×N], wherein N×N signifies the self-attention mechanism through which each token's score is computed by comparing a sentence with itself. Additionally, variables L and H denote the number of layers and self-attention heads, respectively. As evident, it deviates from being a straightforward rank 1 tensor with a size of [N], which contains a score for each token. To tackle this, the package has opted for a simple approach of averaging the scores across layers and heads.

Further experiments using the package highlighted the difficulty in identifying the score differences due to constraints in displaying the results. The rationale behind this is that a 10% alteration in color opacity doesn't bear the same visual significance as it does conceptually, resulting in scores exceeding 70% appearing as very dark red and being attributed greater importance. Put simply, differentiating between words with moderately high or low scores would have been challenging. We observed that employing a basic min-max normalization approach would yield a more uniform outcome.

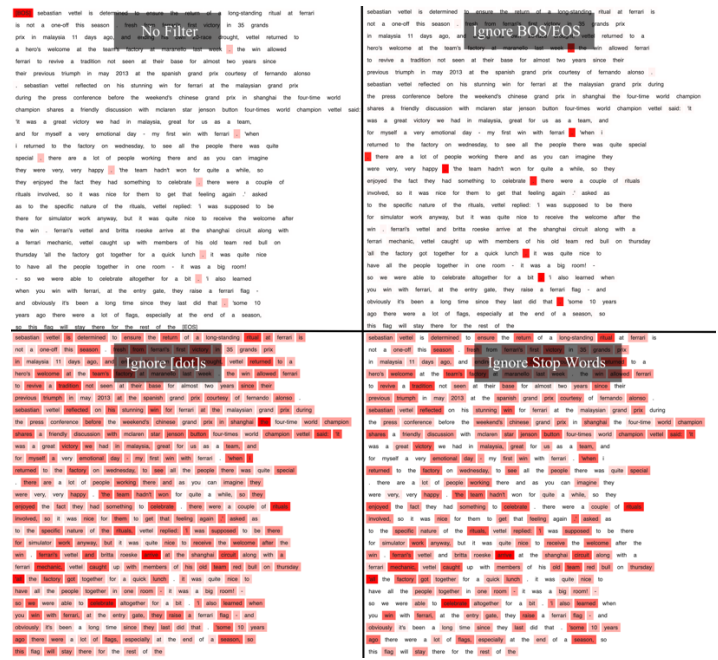

*Figure 2. The impact of distinct filters on the output is showcased through four variations: no filter (top-left), excluding BOS and EOS tokens (top-right), excluding dots (bottom-left), and excluding stop words (bottom-right).*

## 3. Key Features

A limitation of this library might arise from the averaging process, which could potentially lead to information loss when scores are averaged across layers. To address this concern, the controls allow users to examine individual layers/attention heads or group several of them together. It can help with finding patterns in the attention scores and understanding how specific layers or attention heads contribute to the overall representation of the text.

As visible in Figure 2, three options are accessible for enhancing the interpretability of the outcomes. The results are produced by averaging all the attention heads across the layers. The outcomes without any filter place significant emphasis on the special BOS (Beginning of Sentence) and EOS (End of Sentence) tokens. The "Ignore BOS/EOS" option reveals that the dots emerge as the second important element within the texts. Only by masking the scores of dots during calculations can we achieve a more evenly distributed score, allowing us to observe trends effectively. An additional choice has been included to also ignore stop words[1] from the calculation. It might be useful depending on the use case.

It's important to emphasize that when opting for the ignoring options, the tokens are not eliminated from the text. Regardless of the chosen approach, the exact same text will be fed into the RoBERTa model. The sole distinction lies in adjusting the scores of these tokens (after averaging) to the minimum score before undergoing the normalization process.

*Figure 3. An overview of all attention heads in layer 1 of the network. (Appendix 1 contains the images of higher quality.)*

## 4. Usage and Take Aways

The library is designed to work seamlessly with both Jupyter Notebook and, consequently, Google Colab. You can install the package directly from GitHub using PIP, a Python package manager. Subsequently, you can define the visualizer object by initializing the main class. The provided code illustrates sample usage for running an instance. [2]

---

[1] The words such as "a", "are", "is", "the", etc.

[2] The usage of the library is demonstrated in the following URL:
https://colab.research.google.com/github/AlaFalaki/AttentionVisualizer/blob/main/demo.ipynb

```
# Command to install the package
!pip install git+https://github.com/AlaFalaki/AttentionVisualizer.git

# Inializing the UI
import AttentionVisualizer as av
obj = av.AttentionVisualizer()
obj.show_controllers(with_sample=True)
```

*Code 1. The sample code to install and run the package in a Jupyter Notebook environment.*

We can identify several trends by examining the results shown in Figure 3, which display the attention scores of all heads for a single layer. (Refer to Appendix 1 for higher quality images) For example, heads 1 to 7 exhibit an increasing concentration on specific sets of words, particularly evident in head 7. This stands in contrast to the initial group of heads, which distribute attention across a larger number of words. Additionally, it's apparent that head number 8 is specifically trained to focus on hyphenated words such as "one-off", "long-standing", "20-race", etc.

Attention head number 10 views all words in the text as significant. This could potentially serve as a mechanism within the model to transfer information from the initial layers to the later ones. Remarkably, the final head appears to focus on names. These could encompass names of individuals such as "Sebastian", countries like "Malaysia", events like "Prix", or even the names of weekdays such as "Thursday".

## 5. Conclusion

This report delves into the significance of exploring novel approaches for visualizing models. We introduced the "Attention Visualizer" package and made the code accessible by publishing it on GitHub. The details and design decisions involved in implementation of the library are elaborated comprehensively, along with potential enhancements for future versions, including refining the methodology for calculating attention scores. We also conducted a basic analysis to demonstrate the effectiveness of visualization by examining attention patterns within a layer of the RoBERTa model. The illustrations assisted us in pinpointing particular attention heads that focus on patterns like hyphenated words and names. It's evident that this tool has the potential to significantly aid researchers in delving into models with greater detail, thereby enhancing their understanding of the decision-making process and potential biases.

# Appendix 1. Attention Heads Details

In this section, we showcase the high-quality images from Figure 3.

*Figure 4. Attention head number 1 from the first layer.*

sebastian vettel is determined to ensure the return of a long-standing ritual at ferrari is not a one-off this season . fresh from ferrari's first victory in 35 grands prix in malaysia 11 days ago, and ending his own 20-race drought, vettel returned to a hero's welcome at the team's factory at maranello last week . the win allowed ferrari to revive a tradition not seen at their base for almost two years since their previous triumph in may 2013 at the spanish grand prix courtesy of fernando alonso . sebastian vettel reflected on his stunning win for ferrari at the malaysian grand prix during the press conference before the weekend's chinese grand prix in shanghai the four-time world champion shares a friendly discussion with mclaren star jenson button four-times world champion vettel said: 'it was a great victory we had in malaysia, great for us as a team, and for myself a very emotional day - my first win with ferrari . 'when i returned to the factory on wednesday, to see all the people there was quite special . there are a lot of people working there and as you can imagine they were very, very happy . 'the team hadn't won for quite a while, so they enjoyed the fact they had something to celebrate . there were a couple of rituals involved, so it was nice for them to get that feeling again .' asked as to the specific nature of the rituals, vettel replied: 'i was supposed to be there for simulator work anyway, but it was quite nice to receive the welcome after the win . ferrari's vettel and britta roeske arrive at the shanghai circuit along with a ferrari mechanic, vettel caught up with members of his old team red bull on thursday 'all the factory got together for a quick lunch . it was quite nice to have all the people together in one room - it was a big room! - so we were able to celebrate altogether for a bit . 'i also learned when you win with ferrari, at the entry gate, they raise a ferrari flag - and obviously it's been a long time since they last did that . 'some 10 years ago there were a lot of flags, especially at the end of a season, so this flag will stay there for the rest of the

*Figure 5. Attention head number 2 from the first layer.*

sebastian vettel is determined to ensure the return of a long-standing ritual at ferrari is not a one-off this season . fresh from ferrari's first victory in 35 grands prix in malaysia 11 days ago, and ending his own 20-race drought, vettel returned to a hero's welcome at the team's factory at maranello last week . the win allowed ferrari to revive a tradition not seen at their base for almost two years since their previous triumph in may 2013 at the spanish grand prix courtesy of fernando alonso . sebastian vettel reflected on his stunning win for ferrari at the malaysian grand prix during the press conference before the weekend's chinese grand prix in shanghai the four-time world champion shares a friendly discussion with mclaren star jenson button four-times world champion vettel said: 'it was a great victory we had in malaysia, great for us as a team, and for myself a very emotional day - my first win with ferrari . 'when i returned to the factory on wednesday, to see all the people there was quite special . there are a lot of people working there and as you can imagine they were very, very happy . 'the team hadn't won for quite a while, so they enjoyed the fact they had something to celebrate . there were a couple of rituals involved, so it was nice for them to get that feeling again .' asked as to the specific nature of the rituals, vettel replied: 'i was supposed to be there for simulator work anyway, but it was quite nice to receive the welcome after the win . ferrari's vettel and britta roeske arrive at the shanghai circuit along with a ferrari mechanic, vettel caught up with members of his old team red bull on thursday 'all the factory got together for a quick lunch . it was quite nice to have all the people together in one room - it was a big room! - so we were able to celebrate altogether for a bit . 'i also learned when you win with ferrari, at the entry gate, they raise a ferrari flag - and obviously it's been a long time since they last did that . 'some 10 years ago there were a lot of flags, especially at the end of a season, so this flag will stay there for the rest of the

*Figure 6. Attention head number 3 from the first layer.*

sebastian vettel is determined to ensure the return of a long-standing ritual at ferrari is not a one-off this season . fresh from ferrari's first victory in 35 grands prix in malaysia 11 days ago, and ending his own 20-race drought, vettel returned to a hero's welcome at the team's factory at maranello last week . the win allowed ferrari to revive a tradition not seen at their base for almost two years since their previous triumph in may 2013 at the spanish grand prix courtesy of fernando alonso . sebastian vettel reflected on his stunning win for ferrari at the malaysian grand prix during the press conference before the weekend's chinese grand prix in shanghai the four-time world champion shares a friendly discussion with mclaren star jenson button four-times world champion vettel said: 'it was a great victory we had in malaysia, great for us as a team, and for myself a very emotional day - my first win with ferrari . 'when i returned to the factory on wednesday, to see all the people there was quite special . there are a lot of people working there and as you can imagine they were very, very happy . 'the team hadn't won for quite a while, so they enjoyed the fact they had something to celebrate . there were a couple of rituals involved, so it was nice for them to get that feeling again .' asked as to the specific nature of the rituals, vettel replied: 'i was supposed to be there for simulator work anyway, but it was quite nice to receive the welcome after the win . ferrari's vettel and britta roeske arrive at the shanghai circuit along with a ferrari mechanic, vettel caught up with members of his old team red bull on thursday 'all the factory got together for a quick lunch . it was quite nice to have all the people together in one room - it was a big room! - so we were able to celebrate altogether for a bit . 'i also learned when you win with ferrari, at the entry gate, they raise a ferrari flag - and obviously it's been a long time since they last did that . 'some 10 years ago there were a lot of flags, especially at the end of a season, so this flag will stay there for the rest of the

*Figure 7. Attention head number 4 from the first layer.*

sebastian vettel is determined to ensure the return of a long-standing ritual at ferrari is not a one-off this season . fresh from ferrari's first victory in 35 grands prix in malaysia 11 days ago, and ending his own 20-race drought, vettel returned to a hero's welcome at the team's factory at maranello last week . the win allowed ferrari to revive a tradition not seen at their base for almost two years since their previous triumph in may 2013 at the spanish grand prix courtesy of fernando alonso . sebastian vettel reflected on his stunning win for ferrari at the malaysian grand prix during the press conference before the weekend's chinese grand prix in shanghai the four-time world champion shares a friendly discussion with mclaren star jenson button four-times world champion vettel said: 'it was a great victory we had in malaysia, great for us as a team, and for myself a very emotional day - my first win with ferrari . 'when i returned to the factory on wednesday, to see all the people there was quite special . there are a lot of people working there and as you can imagine they were very, very happy . 'the team hadn't won for quite a while, so they enjoyed the fact they had something to celebrate . there were a couple of rituals involved, so it was nice for them to get that feeling again .' asked as to the specific nature of the rituals, vettel replied: 'i was supposed to be there for simulator work anyway, but it was quite nice to receive the welcome after the win . ferrari's vettel and britta roeske arrive at the shanghai circuit along with a ferrari mechanic, vettel caught up with members of his old team red bull on thursday 'all the factory got together for a quick lunch . it was quite nice to have all the people together in one room - it was a big room! - so we were able to celebrate altogether for a bit . 'i also learned when you win with ferrari, at the entry gate, they raise a ferrari flag - and obviously it's been a long time since they last did that . 'some 10 years ago there were a lot of flags, especially at the end of a season, so this flag will stay there for the rest of the

*Figure 8. Attention head number 5 from the first layer.*

sebastian vettel is determined to ensure the return of a long-standing ritual at ferrari is not a one-off this season . fresh from ferrari's first victory in 35 grands prix in malaysia 11 days ago, and ending his own 20-race drought, vettel returned to a hero's welcome at the team's factory at maranello last week . the win allowed ferrari to revive a tradition not seen at their base for almost two years since their previous triumph in may 2013 at the spanish grand prix courtesy of fernando alonso . sebastian vettel reflected on his stunning win for ferrari at the malaysian grand prix during the press conference before the weekend's chinese grand prix in shanghai the four-time world champion shares a friendly discussion with mclaren star jenson button four-times world champion vettel said: 'it was a great victory we had in malaysia, great for us as a team, and for myself a very emotional day - my first win with ferrari . 'when i returned to the factory on wednesday, to see all the people there was quite special . there are a lot of people working there and as you can imagine they were very, very happy . 'the team hadn't won for quite a while, so they enjoyed the fact they had something to celebrate . there were a couple of rituals involved, so it was nice for them to get that feeling again .' asked as to the specific nature of the rituals, vettel replied: 'i was supposed to be there for simulator work anyway, but it was quite nice to receive the welcome after the win . ferrari's vettel and britta roeske arrive at the shanghai circuit along with a ferrari mechanic, vettel caught up with members of his old team red bull on thursday 'all the factory got together for a quick lunch . it was quite nice to have all the people together in one room - it was a big room! - so we were able to celebrate altogether for a bit . 'i also learned when you win with ferrari, at the entry gate, they raise a ferrari flag - and obviously it's been a long time since they last did that . 'some 10 years ago there were a lot of flags, especially at the end of a season, so this flag will stay there for the rest of the

*Figure 9. Attention head number 6 from the first layer.*

sebastian vettel is determined to ensure the return of a long-standing ritual at ferrari is not a one-off this season . fresh from ferrari's first victory in 35 grands prix in malaysia 11 days ago, and ending his own 20-race drought, vettel returned to a hero's welcome at the team's factory at maranello last week . the win allowed ferrari to revive a tradition not seen at their base for almost two years since their previous triumph in may 2013 at the spanish grand prix courtesy of fernando alonso . sebastian vettel reflected on his stunning win for ferrari at the malaysian grand prix during the press conference before the weekend's chinese grand prix in shanghai the four-time world champion shares a friendly discussion with mclaren star jenson button four-times world champion vettel said: 'it was a great victory we had in malaysia, great for us as a team, and for myself a very emotional day - my first win with ferrari . 'when i returned to the factory on wednesday, to see all the people there was quite special . there are a lot of people working there and as you can imagine they were very, very happy . 'the team hadn't won for quite a while, so they enjoyed the fact they had something to celebrate . there were a couple of rituals involved, so it was nice for them to get that feeling again .' asked as to the specific nature of the rituals, vettel replied: 'i was supposed to be there for simulator work anyway, but it was quite nice to receive the welcome after the win . ferrari's vettel and britta roeske arrive at the shanghai circuit along with a ferrari mechanic, vettel caught up with members of his old team red bull on thursday 'all the factory got together for a quick lunch . it was quite nice to have all the people together in one room - it was a big room! - so we were able to celebrate altogether for a bit . 'i also learned when you win with ferrari, at the entry gate, they raise a ferrari flag - and obviously it's been a long time since they last did that . 'some 10 years ago there were a lot of flags, especially at the end of a season, so this flag will stay there for the rest of the

*Figure 10. Attention head number 7 from the first layer.*

sebastian vettel is determined to ensure the return of a long-standing ritual at ferrari is not a one-off this season . fresh from ferrari's first victory in 35 grands prix in malaysia 11 days ago, and ending his own 20-race drought, vettel returned to a hero's welcome at the team's factory at maranello last week . the win allowed ferrari to revive a tradition not seen at their base for almost two years since their previous triumph in may 2013 at the spanish grand prix courtesy of fernando alonso . sebastian vettel reflected on his stunning win for ferrari at the malaysian grand prix during the press conference before the weekend's chinese grand prix in shanghai the four-time world champion shares a friendly discussion with mclaren star jenson button four-times world champion vettel said: 'it was a great victory we had in malaysia, great for us as a team, and for myself a very emotional day - my first win with ferrari . 'when i returned to the factory on wednesday, to see all the people there was quite special . there are a lot of people working there and as you can imagine they were very, very happy . 'the team hadn't won for quite a while, so they enjoyed the fact they had something to celebrate . there were a couple of rituals involved, so it was nice for them to get that feeling again ,' asked as to the specific nature of the rituals, vettel replied: 'i was supposed to be there for simulator work anyway, but it was quite nice to receive the welcome after the win . ferrari's vettel and britta roeske arrive at the shanghai circuit along with a ferrari mechanic, vettel caught up with members of his old team red bull on thursday 'all the factory got together for a quick lunch . it was quite nice to have all the people together in one room - it was a big room! - so we were able to celebrate altogether for a bit . 'i also learned when you win with ferrari, at the entry gate, they raise a ferrari flag - and obviously it's been a long time since they last did that . 'some 10 years ago there were a lot of flags, especially at the end of a season, so this flag will stay there for the rest of the

*Figure 11. Attention head number 8 from the first layer.*

sebastian vettel is determined to ensure the return of a long-standing ritual at ferrari is not a one-off this season . fresh from ferrari's first victory in 35 grands prix in malaysia 11 days ago, and ending his own 20-race drought, vettel returned to a hero's welcome at the team's factory at maranello last week . the win allowed ferrari to revive a tradition not seen at their base for almost two years since their previous triumph in may 2013 at the spanish grand prix courtesy of fernando alonso . sebastian vettel reflected on his stunning win for ferrari at the malaysian grand prix during the press conference before the weekend's chinese grand prix in shanghai the four-time world champion shares a friendly discussion with mclaren star jenson button four-times world champion vettel said: 'it was a great victory we had in malaysia, great for us as a team, and for myself a very emotional day - my first win with ferrari . 'when i returned to the factory on wednesday, to see all the people there was quite special . there are a lot of people working there and as you can imagine they were very, very happy . 'the team hadn't won for quite a while, so they enjoyed the fact they had something to celebrate . there were a couple of rituals involved, so it was nice for them to get that feeling again ,' asked as to the specific nature of the rituals, vettel replied: 'i was supposed to be there for simulator work anyway, but it was quite nice to receive the welcome after the win . ferrari's vettel and britta roeske arrive at the shanghai circuit along with a ferrari mechanic, vettel caught up with members of his old team red bull on thursday 'all the factory got together for a quick lunch . it was quite nice to have all the people together in one room - it was a big room! - so we were able to celebrate altogether for a bit . 'i also learned when you win with ferrari, at the entry gate, they raise a ferrari flag - and obviously it's been a long time since they last did that . 'some 10 years ago there were a lot of flags, especially at the end of a season, so this flag will stay there for the rest of the

*Figure 12. Attention head number 9 from the first layer.*

*Figure 13. Attention head number 10 from the first layer.*

*Figure 14. Attention head number 11 from the first layer.*

sebastian vettel is determined to ensure the return of a long-standing ritual at ferrari is not a one-off this season . fresh from ferrari's first victory in 35 grands prix in malaysia 11 days ago, and ending his own 20-race drought, vettel returned to a hero's welcome at the team's factory at maranello last week . the win allowed ferrari to revive a tradition not seen at their base for almost two years since their previous triumph in may 2013 at the spanish grand prix courtesy of fernando alonso . sebastian vettel reflected on his stunning win for ferrari at the malaysian grand prix during the press conference before the weekend's chinese grand prix in shanghai . the four-time world champion shares a friendly discussion with mclaren star jenson button four-times world champion vettel said: 'it was a great victory we had in malaysia, great for us as a team, and for myself a very emotional day - my first win with ferrari . 'when i returned to the factory on wednesday, to see all the people there was quite special . there are a lot of people working there and as you can imagine they were very, very happy . 'the team hadn't won for quite a while, so they enjoyed the fact they had something to celebrate . there were a couple of rituals involved, so it was nice for them to get that feeling again .' asked as to the specific nature of the rituals, vettel replied: 'i was supposed to be there for simulator work anyway, but it was quite nice to receive the welcome after the win . ferrari's vettel and britta roeske arrive at the shanghai circuit along with a ferrari mechanic, vettel caught up with members of his old team red bull on thursday 'all the factory got together for a quick lunch . it was quite nice to have all the people together in one room - it was a big room! - so we were able to celebrate altogether for a bit . 'i also learned when you win with ferrari, at the entry gate, they raise a ferrari flag - and obviously it's been a long time since they last did that . 'some 10 years ago there were a lot of flags, especially at the end of a season, so this flag will stay there for the rest of the

*Figure 15. Attention head number 12 from the first layer.*